%% file: main.tex
\documentclass[letterpaper, 10 pt, conference]{ieeeconf}  % Comment this line out if you need a4paper

\IEEEoverridecommandlockouts                              % This command is only needed if 
\usepackage{amsmath,amssymb,amsfonts}
\usepackage{bm}

\usepackage{url}
\usepackage{graphics}
\usepackage{algorithmic}
\usepackage{subfigure}
\usepackage{multirow}
\usepackage{booktabs}
\usepackage{multirow}
\usepackage[ruled,linesnumbered]{algorithm2e}
\usepackage{textcomp}
\usepackage{tabularx}
\usepackage{relsize}
\usepackage[flushleft]{threeparttable}
\usepackage{cite}
\usepackage[table,dvipsnames]{xcolor}
\usepackage{colortbl}
\usepackage{enumerate}
\let \labelindent \relax 
\usepackage[inline]{enumitem}
\usepackage{acronym}
\usepackage{siunitx}
\usepackage{comment}
\usepackage{caption}
\usepackage{wrapfig}
\usepackage{graphicx}
\usepackage{balance}
\usepackage{footnote}
\usepackage{tikz}
\usetikzlibrary{positioning, calc, fit, arrows.meta, backgrounds, decorations.pathreplacing, shapes.geometric}
\usepackage{standalone}
\usepackage{amsmath, amssymb}

\usepackage{tikzducks}
\usepackage[acronym,nomain]{glossaries}
\glsdisablehyper

\IEEEaftertitletext{\input{sections/hero_placeholder}\vspace{8pt}}

\makeatletter
\let\NAT@parse\undefined
\renewcommand{\thefigure}{\arabic{section}.\arabic{figure}}
\makeatother
\usepackage{cite} 
\usepackage{hyperref} 

\makeatletter
\renewcommand{\thefigure}{\arabic{figure}}
\makeatother

\title{\LARGE \bf
It Takes Two to Tangram:\\Multi-Agent Object Centric Reinforcement Learning for Assembly
}

\author{
    Paul Knaußt$^{1}$, Kay Pompetzki$^{1}$, Nico Bohlinger$^{1}$, Niklas Funk$^{1}$, Cristiana de Farias$^{1}$,  Rolf Findeisen$^{5}$, Jan Peters$^{1234}$% <-this % stops a space
    \thanks{
    $^{1}$Intelligent Autonomous Systems Lab, Technical University of Darmstadt, Germany.
    $^{2}$Hessian.AI, Germany.
    $^{3}$German Research Center for AI (DFKI), SAIROL, Germany.
    $^{4}$Robotics Institute Germany (RIG).
    $^{5}$Control and Cyber-Physical Systems Laboratory, Technical University of Darmstadt
    \hspace*{1em} Corresponding authors: {\tt\small \{kay, cristiana\}@robot-learning.de }
    }
    \thanks{
    This work was supported by the German Federal Ministry of Research, Technology and Space (BMFTR) under the Robotics Institute Germany (RIG) and the DEMETER project (Grant no.: 01DR25003).
    This work was also supported by a hardware donation from NVIDIA through the Academic Grant Program, and by the Lichtenberg high-performance computer of TU Darmstadt. 
    }
    \thanks{
    Claude (Anthropic) was used for grammar polishing and for coding assistance in the experiments (Section~IV) and in generating the splash images in Fig.~1. All outputs were reviewed and verified by the authors.
    }
    \thanks{This work has been submitted to the IEEE for possible publication. Copyright may be transferred without notice, after which this version may no longer be accessible.}
}

\makeatletter
\let\st@red@title\@title
\def\@title{%
  \begin{flushleft}
    \vspace{-2.0em}
    {\normalfont\small\textit{Preprint}\par}
    \vspace{-10pt}\noindent\rule{\textwidth}{0.1pt}
  \end{flushleft}
  \vskip0.5em\st@red@title}
\makeatother

\begin{document}

\sloppy

%% ---------------------------------- Start ------------------------------ %%

\maketitle\thispagestyle{empty}
\pagestyle{empty}
%%%%%%%%%%%%%%%%%%%%%%%%%%%%%%%%%%%%%%%%%%%%%%%%%%%%%%%%%%%%%%%%%%%%%%%%%%%%%%%%
\input{acronyms}
\input{sections/0_abstract}

\input{sections/1_introduction}
\input{sections/2_related_works}

\input{sections/3_methodology}

\input{sections/4_experiments}
\input{sections/5_conclusion}

\bibliographystyle{IEEEtran}
\bibliography{references_tidy}
\input{sections/6_appendix}

\end{document}

%% file: sections/hero_placeholder.tex
\begin{minipage}{\textwidth}
\normalfont
\centering
\includegraphics[width=0.98\linewidth]{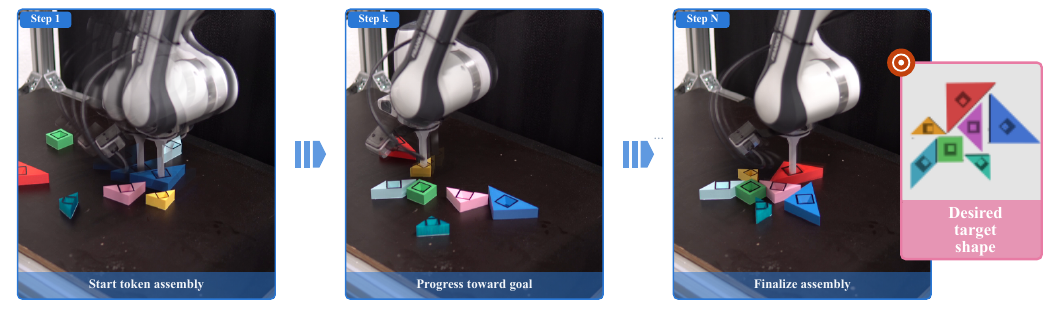}
\captionof{figure}{\acrshort{method} solving the tangram puzzle on our TangramBench.
Given the target assembly (right), we focus on the complex task of coordinating feasible object motions constrained to the support plane. 
\gls{method} predicts object motions that avoid deadlocks and collisions under varying clearance to achieve the final assembly.
}
\label{fig:hero}
\end{minipage}

%% file: acronyms.tex
\newacronym{method}{GOMA}{Graph networks for Object-centric Multi-agent Assembly}
\newacronym{tamp}{TAMP}{task and motion planning}
\newacronym{rl}{RL}{reinforcement learning}
\newacronym{gnn}{GNN}{graph neural network}
\newacronym{ppo}{PPO}{proximal policy optimization}
\newacronym{mappo}{MAPPO}{Multi-Agent Proximal Policy Optimization}
\newacronym{mlp}{MLP}{multilayer perceptron}
\newacronym{gae}{GAE}{generalized advantage estimation}
\newacronym{gelu}{GELU}{Gaussian error linear unit}
\newacronym{rrt}{RRT}{rapidly-exploring random tree}
\newacronym{apf}{APF}{artifical potential field}
\newacronym{pd}{PD}{proportional--derivative}
\newcommand{\bench}{TangramBench\xspace}
\newcommand{\benchT}{TangramBench-Trio\xspace}
\newcommand{\benchC}{TangramBench-Cluster\xspace}
\newcommand{\benchP}{TangramBench-Puzzle\xspace}

%% file: sections/0_abstract.tex
\begin{abstract}
Geometric assembly requires fine spatiotemporal coordination to place parts without causing collisions or deadlocks under tight clearances.
To address this, we present an object-centric \glsentrylong{rl} framework that treats each part as an agent and learns a shared policy for continuous assembly control.
A \glsentrylong{gnn} represents the pieces, their target poses, and their spatial relationships, and predicts translational and angular velocities for each component.
Per-agent \glsentrylong{ppo} with a centralized graph critic directly learns motion and temporal coordination within a single feedback policy.
We study this formulation through planar Tangram assembly, where pieces must slide and rotate into a given configuration while remaining on the assembly surface.
Procedural generation of target configurations and randomized initial arrangements support training and evaluation across diverse assembly problems.
Our experiments isolate the roles of relational structure and multi-agent optimization, and compare learned control with sampling-based planning.
We demonstrate concurrent assembly in simulation and transfer the policy to sequential execution on a real robot, as our object-centric formulation decouples learned assembly coordination from low-level physical control.
\textit{Project page}: https://tangrambench.org/
\end{abstract}

%% file: sections/1_introduction.tex
\section{Introduction}
\label{sec:introduction}

From arranging or assembling furniture in a home to fitting components together on a production line, many tasks require moving multiple objects into precise spatial relationships, often within cluttered or confined spaces~\cite{knepper2013ikeabot, kirschner2016yumi, johns2023framework}.
In such cases, even when every object's destination is known, its motion cannot be considered in isolation.
To illustrate the problem, consider sliding puzzle pieces across a table to form a given shape.
As one piece approaches its target, it may narrow or block the path of another.
Once the desired configuration is known, the challenge becomes coordinating the pieces' motions toward their targets while accounting for the space they need along the way. 

One approach to the coordination problem is to jointly plan which objects to move and how to move them.
\Gls{tamp} methods couple these decisions by searching for action sequences whose associated motions satisfy geometric and robot constraints~\cite{garrett2021integrated}.
In robot assembly, this involves selecting part sequences and assembly configurations together with feasible manipulation motions~\cite{tian2024asap}.
In nonprehensile rearrangement, object-centric approaches guide planning through actions directed at individual objects~\cite{king2016objectcentric}, or compute desired object motions that are realized through closed-loop pushing~\cite{ren2025objectcentric}.
This abstraction organizes manipulation around changes to the objects whose configurations define the task.
Learning-based approaches also address multi-object rearrangement through hierarchical policies~\cite{bai2022hierarchical}, while assembly methods exploit structured representations of parts and their relationships to select placements~\cite{funk2022learn2assemble, funk2022milp}
or generate object motions~\cite{ghasemipour2022blocks}.
These approaches establish objects and their relationships as useful units for reasoning about manipulation.

Multi-agent pathfinding offers another perspective on coordination, studying how agents can reach their respective goals while avoiding conflicts during concurrent movement~\cite{stern2019mapf}.
Coordinating these movements requires reasoning about interactions between agents and the space they share.
Graph representations provide a way to encode these relationships, supporting learned policies for decentralized path planning~\cite{li2020graph}, multi-agent navigation and collision avoidance~\cite{nayak2023informarl}, and coordinated motion, task allocation, and scheduling in multi-robot reaching~\cite{lai2025roboballet}.
This perspective naturally extends to object-centric assembly and rearrangement, where individual objects can be represented as agents acting toward a shared target configuration.

Thus, we introduce \gls{method}, a graph-based multi-agent approach to geometric assembly.
\glsentryshort{method} represents each part to be assembled as a cooperative agent and uses a shared policy trained with \gls{mappo}~\cite{yu2022mappo} to predict continuous object motions from a relational representation of the assembly state.
A \gls{gnn} encodes interactions between parts, combining an object-centric representation with multi-agent learning toward a shared assembly objective.
Our central question is whether an object-centric multi-agent formulation provides an effective basis for learning coordinated assembly.
We further investigate which aspects of the representation and learning formulation matter for coordination, including how information about other parts informs each agent's actions.

To study this problem, we also introduce \bench, a procedural benchmark suite for planar geometric assembly based on the game of Tangram.
The classical Tangram puzzle consists of seven geometric pieces (see Fig.~\ref{fig:hero}) that must be arranged to form a target silhouette.
Despite its small set of simple parts, tightly packed configurations create strong geometric dependencies between their motions, making Tangram a compact setting for studying coordinated assembly.
Tangram has previously served as a robot testbed for collaborative pick-and-place~\cite{kirschner2016yumi} and sequential assembly from silhouette prompts~\cite{zhao2025mrchaos}, while recent benchmarks study geometric reasoning over valid piece configurations~\cite{zong2026tangramsr,liu2026tangrampuzzle}.
In contrast, \bench provides the target pose of each piece and constrains its motion to the support plane, isolating the problem of coordinating feasible object motions toward a known assembly.
We procedurally generate three levels of geometric difficulty, from loosely packed configurations to compact assemblies in which interactions between pieces become increasingly important.
The benchmark supports large-scale training and evaluation and includes simulation-ready models together with physical pieces and a compatible manipulation tool for real-robot experiments.

In summary, our contributions are:
\begin{itemize}[leftmargin=*,nosep]
\item An object-level multi-agent formulation of geometric assembly that learns simultaneous part motions and temporal coordination through a shared continuous-control policy.
\item An empirical investigation of relational state representations and per-agent policy optimization, with comparisons against joint-action learning and sampling-based planning.
\item \bench, a procedural benchmark suite for geometric assembly, with physical pieces and evaluations of concurrent simulation control and sequential robot execution.
\end{itemize}

%% file: sections/2_related_works.tex
\section{Related Work}
\label{sec:relatedworks}
\subsection{Assembly and Motion Planning}
Task and motion planning couples discrete task decisions with continuous geometric constraints~\cite{garrett2021integrated}.
Assembly planners exploit disassembly to recover feasible sequences and motions, accounting for part geometry, stability, and graspability~\cite{tian2022assemble,tian2024asap}.
Fabrica extends this approach with learned controllers for contact-rich execution~\cite{tian2025fabrica}, while~\cite{hartmann2023construction} coordinate concurrent construction through optimization and space-time search over smaller task subproblems.
These approaches explicitly search for feasible assembly operations and their execution.
Additionally, sampling-based methods such as RRT-Connect~\cite{kuffner2000rrtconnect} 
and Global Tensor Motion Planning~\cite{le2025gtmp} % uncommented
construct paths through collision-free configurations.
The latter vectorizes sampling, collision checking, and search. % uncommented
Learned state and action proposals can guide physics-based rearrangement planning~\cite{haustein2019rearrangement}, and learned tree-search policies can select objects and pushing primitives~\cite{bai2022hierarchical}.
Hierarchical policy blending combines reactive control with planning over the weights of expert policies~\cite{hansel2023blending, le2023blending}.
Our approach instead learns a feedback policy that coordinates part velocities from the assembly state without an online search over object trajectories, manipulation primitives, or controller weights.

\subsection{Learning for Assembly and Rearrangement}
Several methods have applied learning algorithms to assembly tasks. Learn2Assemble 
and its extension with mixed-integer programming 
combines structured \glsentrylong{rl} and search with geometric and robot feasibility~\cite{funk2022learn2assemble, funk2022milp}.
Other methods learn feasible assembly sequences~\cite{shu2024subassembly} or alternative plans under execution disruptions~\cite{wang2025alternative}. Geometric assembly methods also infer the desired part arrangement.
Dynamic graph learning predicts part poses from geometric input~\cite{huang2020assembly}.
Particularly for geometric assembly, MRChaos learns visual Tangram assembly through sequential pick-and-place actions in a pre-defined part order~\cite{zhao2025mrchaos}.
Blocks Assemble!~\cite{ghasemipour2022blocks} uses a graph policy to predict block velocities and select blocks for two virtual grippers, with large-scale \gls{ppo} training across diverse blueprints.
Kataoka et al.~\cite{kataoka2023bimanual} transfer joint-space \glsentrylong{rl} to bimanual magnetic block assembly.
Conversely, our approach uses object relations to coordinate continuous part velocities, coupling movement and temporal ordering throughout execution rather than only when selecting operations. Here we take assigned target poses as input and learn the planar motions. During execution, pieces remain on the assembly surface so that their shapes and poses constrain one another. Overall, our approach differs from previous geometric assembly works as we learn the object-level velocities with a multi-agent learning formulation, assigning a policy output and optimization objective to each piece while retaining a shared graph representation that supports both concurrent assembly in simulation and sequential execution with a real robot.

\subsection{Graph Policies and Multi-Agent Coordination}
Coordinating multiple agents requires accounting for how each agent's motion affects the others. Multi-agent pathfinding studies this problem through interacting routes and temporal collision constraints~\cite{stern2019mapf}. In learned policies, graph networks provide a natural way to represent these interactions through shared relational computations~\cite{battaglia2018relational}. InforMARL~\cite{nayak2023informarl}, for example, combines graph-based information aggregation with \gls{mappo}~\cite{yu2022mappo} to learn navigation policies with centralized value estimation. RoboBallet~\cite{lai2025roboballet} applies graph-based coordination to multiple robot arms, jointly addressing task allocation, scheduling, and motion. We adopt a similar perspective for assembly, treating the pieces themselves as cooperative agents. Here, our approach moves the coordination problem from robot arms to assembly pieces, using a shared graph policy to learn coupled motions whose feasibility depends on the shapes and poses of the other parts as they form the final structure.

\subsection{Robot Assembly Benchmarks}

Assembly benchmarks span tasks ranging from determining valid part configurations to executing the manipulation required to realize them.
Tangram provides a particularly compact instance of this problem and has been used across several settings.
Kirschner et al.~\cite{kirschner2016yumi} use a five-piece Tangram for collaborative robot assembly, where pieces are grasped and placed at taught target locations, while Zhao et al.~\cite{zhao2025mrchaos} learn sequential pick-and-place assembly of the seven Tangram pieces from a target silhouette.
Outside physical manipulation, generated Tangram datasets have been used to learn mappings from silhouettes to complete solutions~\cite{yamada2025tangan}, including TANGAN with thousands of randomly generated puzzles~\cite{yamada2025tangan}.

These works primarily evaluate \emph{which configuration} solves a puzzle or the accuracy with which pieces can be placed.
In contrast, \bench takes the desired configuration as given and evaluates the coordinated planar manipulation required to reach it from an initial arrangement, making interactions and conflicts between object motions part of the benchmark itself.

%% file: sections/3_methodology.tex
\section{\bench}
\label{sec:bench}

\begin{figure}[t]
  \centering
  \includegraphics[width=\linewidth]{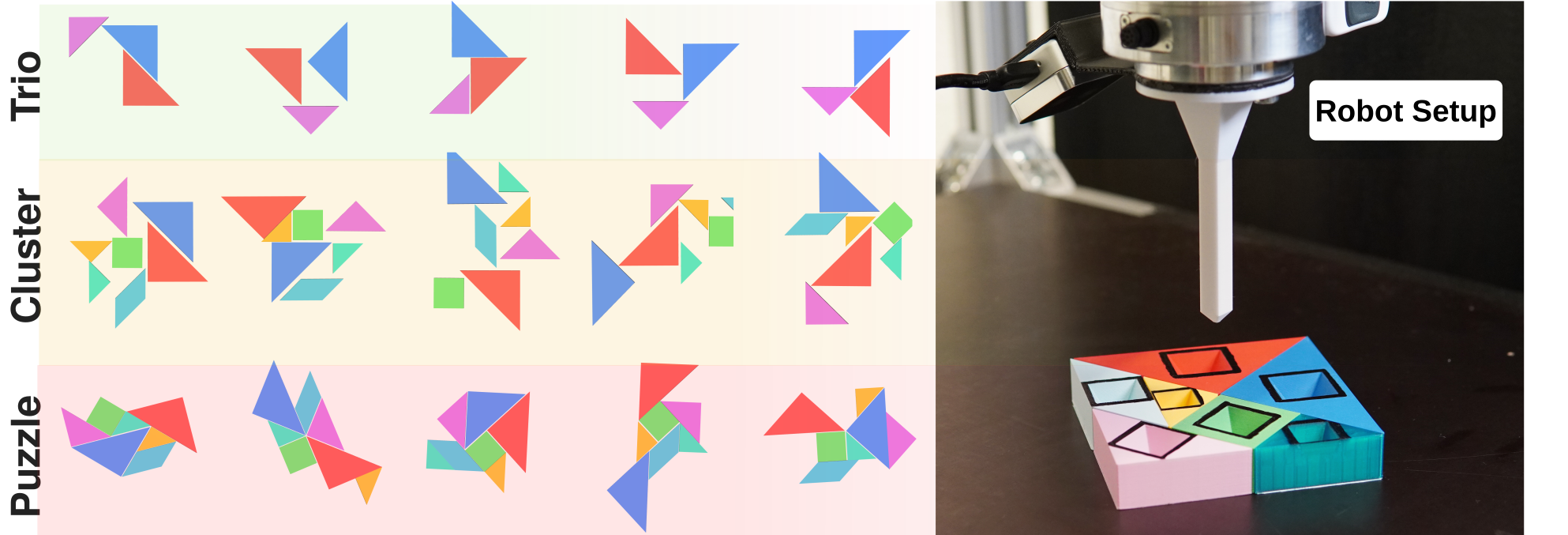}
\caption{\bench provides a graded benchmark for planar geometric assembly, ranging from the three-piece \benchT to seven-piece \benchC and tightly constrained \benchP configurations. We provide simulation-ready models, together with a robot-compatible physical counterpart consisting of 3D-printed pieces and an insertion tool for planar manipulation.}
  \label{fig:benchmark}
\end{figure}

\bench is a procedural benchmark for planar geometric assembly built around the seven pieces of the classical Tangram puzzle. As illustrated in Fig.~\ref{fig:benchmark}, the set consists of two large triangles, one medium triangle, two small triangles, a square, and a parallelogram. Although each individual piece has only three planar degrees of freedom (position and yaw), assembling several pieces into a compact configuration gives rise to a coupled, constrained motion-planning problem. As target slots become occupied, the collision-free space available to the remaining pieces changes and may contain increasingly narrow passages. A motion that is locally valid for one piece can therefore obstruct the route of another, making successful assembly depend not only on reaching the individual targets, but also on coordinating how the pieces approach them.

For \bench we therefore consider a set of $N$ rigid planar pieces, with the configuration of piece $i$ given by
\begin{equation}
q_i = (p_i,\vartheta_i)
\in \mathrm{SE}(2),
\qquad
p_i=(x_i,y_i)\in\mathbb{R}^2 ,
\end{equation}
and denote the joint configuration of all pieces by
$q=(q_1,\ldots,q_N)$.
Each task instance therefore specifies an initial configuration
$q^0$ and an assigned target configuration
$q^\star=(q_1^\star,\ldots,q_N^\star)$.
The objective here is to drive all pieces from $q^0$ to their corresponding targets while keeping them within the support plane and avoiding inter-piece penetration throughout the motion. 

Because the piece-to-target assignment is provided, \bench deliberately removes assignment and shape-matching ambiguities and isolates the coordination problem. That is, \bench central question is: 
\begin{center}
    \emph{How should the pieces move to their known targets without preventing one another from reaching them?}
\end{center}

We show in Fig.~\ref{fig:benchmark} an overview of our benchmark suite. For each instance, target poses $q_i^\star=(p_i^\star,\vartheta_i^\star)$ are placed on a $70\times36$cm support plane. Initial poses are sampled within a $60\times34$cm region, with piece centers separated by at least $18$cm. Pieces may translate and rotate freely on the plane. We consider a piece successfully placed when its center is within $1$cm of the target position and with an orientation error below $0.1$rad. To expose different degrees of geometric coupling, \bench contains three levels of increasing difficulty. (i) \textbf{\benchT} contains only three pieces---the two large triangles and the medium triangle---and provides a reduced setting in which coordination can be studied without the constraints induced by the full puzzle. (ii) \textbf{\benchC} uses all seven pieces and generates compact target configurations procedurally. Pieces are first scattered and then attracted toward a common center until contact prevents further motion, with configurations containing interpenetrations rejected. This process produces irregular but relatively open assemblies, for which most target slots remain accessible from outside the configuration. Finally, (iii) \textbf{\benchP} contains tightly packed Tangram configurations constructed by joining pieces edge-to-edge at compatible orientations, making them closer to classical Tangram figures. These layouts create enclosed target slots and narrow access corridors, such that placing one piece can remove feasible approaches to another and the order of insertion becomes consequential. Despite having the same median inter-piece spacing of $37$mm, \benchC instances have a median enclosure of only $9\%$, compared with $57\%$ for \benchP instances. This increased enclosure makes \benchP substantially more constrained, as feasible approach directions are reduced and successful assembly depends more strongly on coordinated sequencing of the pieces. All configurations are generated procedurally from fixed seeds, allowing disjoint training and evaluation pools of arbitrary size to be reproduced. Here, we provide approximately $1$M, $1.1$M, and $955$k training configurations for \benchT, Cluster, and Puzzle, respectively. Additionally, we provide $258$ held-out configurations for each evaluation\footnote{The camera-ready version will link the repository containing the generators, assets, and benchmark splits.}.

For evaluation, we provide a MuJoCo~\cite{todorov2012mujoco} implementation of \bench with mesh-based collision between pieces. Following the object-centric formulation of the benchmark, each simulated piece is directly actuated in its planar degrees of freedom. Velocity actuators realize commanded body-frame translational and angular velocities, allowing algorithms to focus on coordination between objects rather than on the kinematics of a particular robot platform. Finally, \bench provides a common set of evaluation metrics, including assembly success, position and yaw error, inter-piece contact rate, task completion time, and computation time. We also report the fraction of pieces within $5$cm of their assigned targets. To support evaluation on physical hardware, we additionally provide a real-world counterpart using 3D-printed pieces with footprints matching the simulated models. Each piece contains a tapered insertion socket at its centroid that can be engaged by a custom-designed, square-shaped peg manipulation tool with a chamfered tip mounted to the robot (see in Fig. \ref{fig:benchmark}). This interface intentionally abstracts away grasp acquisition and constrains the interaction to planar translation and rotation, preserving the object-level planning problem studied in simulation while reducing the sim-to-real gap. The robot can therefore move one piece at a time using the same planar commands produced by the object-centric policy.

\section{\glsentrylong{method} (\glsentryshort{method})}
\label{sec:method}

\begin{figure*}[t]
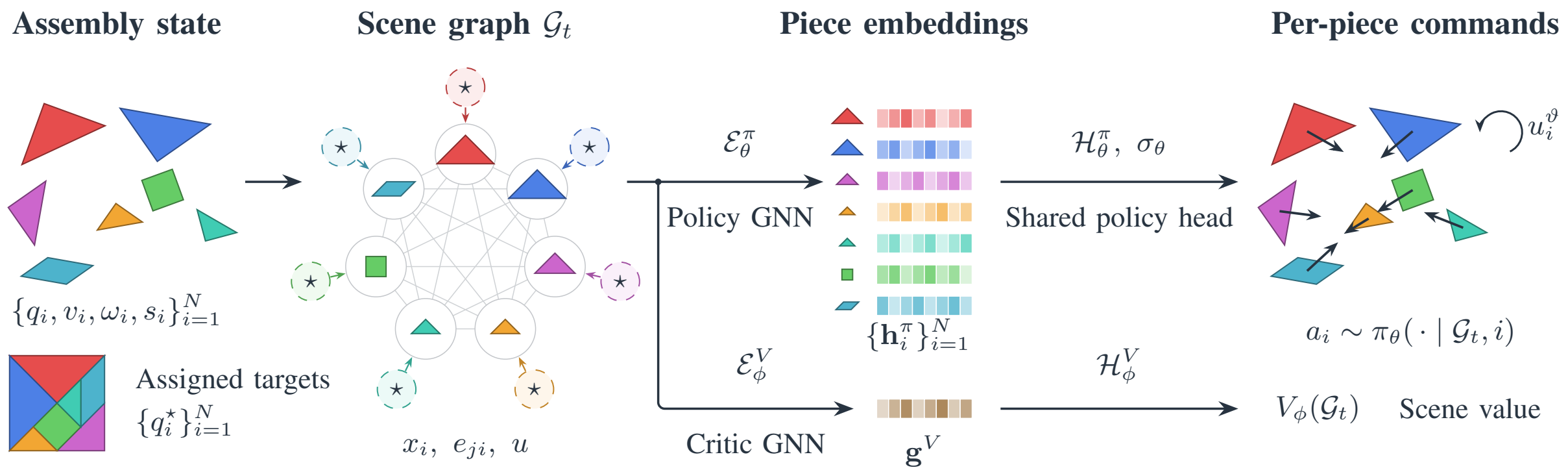

  \centering
  \includestandalone[width=\textwidth]{figures/architecture}
  \caption{\glsentryshort{method} maps a graph of pieces and assigned target poses to per-piece velocity commands.
  Independent graph encoders produce piece embeddings for a shared policy head and a global embedding for the critic.}
  \label{fig:arch}
\end{figure*}

Here, we provide an overview of \gls{method}, seen in Fig.~\ref{fig:arch}.

\subsection{Policy Architecture}
\label{sec:architecture}

We use separate policy and critic \glspl{gnn}, following the edge, node, and global update structure of graph networks~\cite{battaglia2018relational}.
We denote the encoders by $\mathcal{E}^{\pi}_{\theta}$ and $\mathcal{E}^{V}_{\phi}$, with policy parameters $\theta$ and critic parameters $\phi$.
Within either encoder, learned embedding maps 
$\mathcal{B}_v$, $\mathcal{B}_e$, and $\mathcal{B}_g$
project raw node features $x_i$, edge features $e_{ji}$, and global features $u$ to dimension $H$, which gives the initial node, edge, and global embeddings
\begin{equation}
\mathbf{h}_i^{(0)}=\mathcal{B}_v(x_i),\quad
\mathbf{z}_{ji}^{(0)}=\mathcal{B}_e(e_{ji}),\quad
\mathbf{g}^{(0)}=\mathcal{B}_g(u).
\label{eq:graph_embeddings}
\end{equation}
Embedding and update maps are residual \glspl{mlp}.

Each message-passing round updates edges from their previous embeddings, the sender and receiver embeddings, and the global context.
Node updates combine the current node embedding with sums of incoming and outgoing edge embeddings, and the global update aggregates the updated nodes and edges.
Writing $\mathcal{F}_e$, $\mathcal{F}_v$, and $\mathcal{F}_g$ for these maps and $[\cdot]$ for concatenation, a message-passing round $k$ computes
\begin{equation}
\begin{aligned}
\mathbf{z}_{ji}^{(k+1)}&=\mathcal{F}_e\big([\mathbf{z}_{ji}^{(k)},\mathbf{h}_j^{(k)},\mathbf{h}_i^{(k)},\mathbf{g}^{(k)}]\big),\\
\mathbf{m}_{i,\mathrm{in}}^{(k+1)}&=\textstyle\sum_{j\rightarrow i}\mathbf{z}_{ji}^{(k+1)},\qquad
\mathbf{m}_{i,\mathrm{out}}^{(k+1)}=\sum_{i\rightarrow j}\mathbf{z}_{ij}^{(k+1)},\\
\mathbf{h}_i^{(k+1)}&=\textstyle\mathcal{F}_v\big([\mathbf{h}_i^{(k)},\mathbf{m}_{i,\mathrm{in}}^{(k+1)},\mathbf{m}_{i,\mathrm{out}}^{(k+1)},\mathbf{g}^{(k)}]\big),\\
\mathbf{g}^{(k+1)}&=\textstyle\mathcal{F}_g\Big(\Big[\mathbf{g}^{(k)},\sum_i\mathbf{h}_i^{(k+1)},\sum_{j\rightarrow i}\mathbf{z}_{ji}^{(k+1)}\Big]\Big).
\end{aligned}
\label{eq:message_passing}
\end{equation}
The global sums include all nodes and edges in $\mathcal{G}_t$, and each update map is shared across entities of the corresponding type.

Let $\mathbf{h}^{\pi}_i$ be the final piece embedding from $\mathcal{E}^{\pi}_{\theta}$ and $\mathbf{g}^{V}$ the final global embedding from $\mathcal{E}^{V}_{\phi}$.
The policy head $\mathcal{H}^{\pi}_{\theta}$ is shared across pieces, while the critic head $\mathcal{H}^{V}_{\phi}$ outputs one scalar for the scene.
With action scaling factors $a_{\max}$, the network outputs are
\begin{equation}
\begin{aligned}
\mu_{\theta,i}=a_{\max}\odot\tanh\big(\mathcal{H}^{\pi}_{\theta}(\mathbf{h}^{\pi}_i)\big),\quad
V_{\phi}(\mathcal{G})=\mathcal{H}^{V}_{\phi}(\mathbf{g}^{V}).
\end{aligned}
\label{eq:network_heads}
\end{equation}
Each piece uses a diagonal Gaussian action distribution with mean $\mu_{\theta,i}$ and a learned standard deviation vector $\sigma_\theta$ shared across pieces.
Table~\ref{tab:hyperparameters} summarizes our hyperparameters.

\subsection{Policy Training}
\label{sec:training}

We train on parallel simulation episodes with sampled target configurations and randomized initial piece poses.
All pieces are actuated concurrently, with each command held over a fixed interval of simulation steps.
The reward combines progress and pose accuracy for each piece with contact penalties, workspace penalties, and a shared incentive to complete the assembly.
We define the distance error $d_i=\|p_i^\star-p_i\|$ and let $\epsilon_i$ be the wrapped orientation error.
With that, we define $I_i$ as the completion indicator that $d_i<\tau_p$ and $|\epsilon_i|<\tau_\vartheta$.
We use $\tau_p=0.01\,\mathrm{m}$ and $\tau_\vartheta=0.1\,\mathrm{rad}$ as the completion thresholds.
Based on this, the individual reward is defined as
\begin{equation}
\begin{aligned}
\textstyle r_i^{\mathrm{ind}}={}&w_v\frac{(p_i^\star-p_i)^\top v_i}{d_i+\varepsilon}
\frac{\|v_i\|}{v_{\mathrm{ref}}}
-w_\vartheta(1-\cos\epsilon_i)\\
&+w_s I_i-w_c\textstyle\sum_{j\ne i}c_{ij}-w_b b_i.
\end{aligned}
\label{eq:individual_reward}
\end{equation}
The progress term projects velocity onto the target direction and scales it by speed relative to $v_{\mathrm{ref}}$.
The contact indicator $c_{ij}$ is accumulated over simulation steps within each control interval, capturing transient contacts.
Contacts do not terminate an episode, allowing the policy to correct configurations after an interaction.
The contact cost discourages interference between moving pieces, while the pose reward favors closing the gaps required for compact assemblies.
The workspace term $b_i$ combines positive-part penalties for crossing the rectangular workspace limits and entering a margin near its boundary.
The sparse pose reward is supplied at every step that the piece remains within both target tolerances.

The shared reward penalizes the maximum positional error and rewards completion of all target poses,
\begin{equation}
\textstyle
r_i=r_i^{\mathrm{ind}}-w_d\max_j d_j+w_T\prod_{j=1}^{N}I_j.
\label{eq:team_reward}
\end{equation}
These shared terms keep the reward sensitive to unfinished placements even when a piece has reached its own target.

We optimize the policy with the per-agent clipped objective of following \gls{mappo}~\cite{yu2022mappo} with the reward of Eq. \eqref{eq:team_reward}.
\Gls{gae} computes a separate advantage for each piece from its reward $r_{i,t}$ and the shared scene baseline $V_\phi(\mathcal G_t)$.
Each likelihood ratio covers the three command components of one piece, and the clipped surrogate is averaged over pieces and scene transitions.
This separates the clipping across pieces, whereas joint-action \gls{ppo} would couple them through a single ratio for all piece commands.
The critic predicts one scene value and is trained toward the mean of the per-piece return targets.
This keeps individual rewards in the policy updates while sharing the baseline across pieces.

%% file: figures/architecture.tex
\definecolor{ink}{HTML}{273340}
\definecolor{policy}{HTML}{176FA0}
\definecolor{critic}{HTML}{906025}
\definecolor{pBT1}{rgb}{0.90,0.30,0.30}   % big_triangle_1
\definecolor{pBT2}{rgb}{0.30,0.50,0.90}   % big_triangle_2
\definecolor{pMT} {rgb}{0.80,0.40,0.80}   % middle_triangle
\definecolor{pST1}{rgb}{0.95,0.65,0.20}   % small_triangle_1
\definecolor{pST2}{rgb}{0.25,0.80,0.70}   % small_triangle_2
\definecolor{pSQ} {rgb}{0.40,0.80,0.40}   % square
\definecolor{pPAR}{rgb}{0.30,0.70,0.80}   % parallelogram

\begin{tikzpicture}[
  font=\fontfamily{ptm}\fontsize{9}{11}\selectfont,text=ink,>=Stealth,
  flow/.style={->,draw=ink,line width=.7pt,rounded corners=2pt},
  title/.style={font=\fontfamily{ptm}\fontsize{10}{12}\selectfont\bfseries},
  pc/.style={line width=.4pt},
  pA/.style={pc,draw=pBT1!60!black,fill=pBT1},
  pB/.style={pc,draw=pBT2!60!black,fill=pBT2},
  pC/.style={pc,draw=pMT!60!black,fill=pMT},
  pD/.style={pc,draw=pST1!60!black,fill=pST1},
  pE/.style={pc,draw=pST2!60!black,fill=pST2},
  pF/.style={pc,draw=pSQ!60!black,fill=pSQ},
  pG/.style={pc,draw=pPAR!60!black,fill=pPAR},
]
% All pieces preserve the area ratios of a standard seven-piece Tangram.
% Local coordinates place each piece's centroid at the origin.
\newcommand{\tangram}[1]{%
  \ifnum#1=1 \path[pA] (-.5,-.1667)--(.5,-.1667)--(0,.3333)--cycle;\fi
  \ifnum#1=2 \path[pB] (-.5,-.1667)--(.5,-.1667)--(0,.3333)--cycle;\fi
  \ifnum#1=3 \path[pC] (-.3536,-.1179)--(.3536,-.1179)--(0,.2357)--cycle;\fi
  \ifnum#1=4 \path[pD] (-.25,-.0833)--(.25,-.0833)--(0,.1667)--cycle;\fi
  \ifnum#1=5 \path[pE] (-.25,-.0833)--(.25,-.0833)--(0,.1667)--cycle;\fi
  \ifnum#1=6 \path[pF] (-.1768,-.1768) rectangle (.1768,.1768);\fi
  \ifnum#1=7 \path[pG] (-.375,-.125)--(.125,-.125)--(.375,.125)--(-.125,.125)--cycle;\fi
}
\newcommand{\tangramSquare}{%
  \path[pA] (-0.7188,0.7188) -- (0.7188,0.7188) -- (0.0000,0.0000) -- cycle;
  \path[pB] (-0.7188,-0.7188) -- (-0.7188,0.7188) -- (0.0000,0.0000) -- cycle;
  \path[pC] (0.0000,-0.7188) -- (0.7188,-0.7188) -- (0.7188,0.0000) -- cycle;
  \path[pF] (-0.3594,-0.3594) -- (0.0000,-0.7188) -- (0.3594,-0.3594) -- (0.0000,0.0000) -- cycle;
  \path[pG] (0.3594,-0.3594) -- (0.7188,0.0000) -- (0.7188,0.7188) -- (0.3594,0.3594) -- cycle;
  \path[pD] (-0.7188,-0.7188) -- (0.0000,-0.7188) -- (-0.3594,-0.3594) -- cycle;
  \path[pE] (0.0000,0.0000) -- (0.3594,-0.3594) -- (0.3594,0.3594) -- cycle;
}

\path[use as bounding box] (-.12,-.05) rectangle (17.65,5.15);
\node[title] at (1.30,4.85) {Assembly state};
\node[title] at (5.10,4.85) {Scene graph $\mathcal G_t$};
\node[title] at (9.88,4.85) {Piece embeddings};
\node[title] at (15.45,4.85) {Per-piece commands};

% A physical scene and its assigned target configuration.
\foreach \i/\x/\y/\r in {1/.65/3.7/25,2/1.8/3.7/170,3/.35/2.83/80,4/1.32/2.75/10,5/2.35/2.65/150,6/1.8/3.06/20,7/.65/2.18/-10}{
  \begin{scope}[shift={(\x,\y)},rotate=\r] \tangram{\i} \end{scope}
}
\node at (1.3,1.75) {$\{q_i,v_i,\omega_i,s_i\}_{i=1}^{N}$};
\begin{scope}[shift={(.65,.74)},scale=.72] \tangramSquare \end{scope}
\node[align=left,anchor=west] at (1.38,.74) {Assigned targets\\[2pt]$\{q_i^\star\}_{i=1}^{N}$};

% Piece and target nodes share the colors of their physical counterparts.
\foreach \i/\a in {1/90,2/38.57,3/-12.86,4/-64.29,5/-115.71,6/-167.14,7/141.43}{
  \coordinate (v\i) at ({5.1+1.00*cos(\a)},{2.45+1.00*sin(\a)});
  \coordinate (t\i) at ({5.1+1.73*cos(\a)},{2.45+1.73*sin(\a)});
}
\foreach \i in {1,...,7}{\foreach \j in {1,...,7}{
  \ifnum\i<\j \draw[draw=ink!22,line width=.3pt,shorten <=3.3mm,shorten >=3.3mm] (v\i)--(v\j);\fi
}}
\foreach \i/\c in {1/pBT1,2/pBT2,3/pMT,4/pST1,5/pST2,6/pSQ,7/pPAR}{
  \node[circle,fill=white,inner sep=0pt,draw=ink!28,line width=.3pt,minimum size=6.6mm] at (v\i) {};
  \begin{scope}[shift={(v\i)},scale=.62] \tangram{\i} \end{scope}
  \node[circle,draw=\c!80!black,dashed,fill=\c!10,inner sep=0pt,minimum size=4.2mm,font=\fontsize{8}{9}\selectfont] at (t\i) {$\star$};
  \draw[->,dashed,draw=\c!80!black,line width=.45pt,shorten <=2.2mm,shorten >=3.4mm,>={Stealth[length=1.1mm]}] (t\i)--(v\i);
}
\draw[flow] (2.7,3.15)--(3.32,3.15);
\node at (5.1,.24) {$x_i,\ e_{ji},\ u$};

% Encoders are learned maps on arrows rather than text-filled boxes.
\draw[draw=ink,line width=.7pt] (6.86,3.15)--(7.2,3.15);
\fill[ink] (7.2,3.15) circle (1pt);
\draw[flow] (7.2,3.15)--node[above=4pt] {$\mathcal E^\pi_\theta$} node[pos=.42,below=4pt,xshift=4pt] {Policy GNN} (9.00,3.15);
\foreach \i/\c in {1/pBT1,2/pBT2,3/pMT,4/pST1,5/pST2,6/pSQ,7/pPAR}{
  \pgfmathsetmacro\yy{4.18-.34*\i}
  \begin{scope}[shift={(9.26,\yy)},scale=.33] \tangram{\i} \end{scope}
  \foreach \j in {0,...,7}{
    \pgfmathtruncatemacro\shade{20+mod(17*\i+23*\j,65)}
    \path[draw=white,line width=.3pt,fill=\c!\shade] ({9.58+.13*\j},{\yy-.105}) rectangle ++(.13,.21);
  }
}
\node at (10.03,1.48) {$\{\mathbf h_i^\pi\}_{i=1}^{N}$};

% A shared head maps each embedding to a separate velocity distribution.
\draw[flow] (10.93,3.15)--node[above=4pt] {$\mathcal H^\pi_\theta,\ \sigma_\theta$} node[below=4pt,align=center] {Shared policy head} (13.52,3.15);
% The output scene is the input scene translated by 13.62 cm on the page.
% Goal offsets are the centroids of tangramSquare, scaled to the piece sizes
% and centered at (1.30, 3.00) in the input scene's coordinate system.
% Every displayed translation arrow follows its piece-to-goal vector.
\foreach \i/\x/\y/\r/\gx/\gy in {1/.65/3.7/25/0/.3333,2/1.8/3.7/170/-.3333/0,3/.35/2.83/80/.3333/-.3333,4/1.32/2.75/10/-.25/-.4167,5/2.35/2.65/150/.1667/0,6/1.8/3.06/20/0/-.25,7/.65/2.18/-10/.375/.125}{
  \pgfmathsetmacro\xx{\x+13.62}
  \pgfmathsetmacro\dx{1.30+\gx-\x}
  \pgfmathsetmacro\dy{3.00+\gy-\y}
  \pgfmathsetmacro\arrowlength{min(.46,veclen(\dx,\dy))}
  \pgfmathsetmacro\arrowangle{atan2(\dy,\dx)}
  \begin{scope}[shift={(\xx,\y)},rotate=\r] \tangram{\i} \end{scope}
  \draw[->,draw=ink,line width=.75pt,>={Stealth[length=1.6mm]}] (\xx,\y)--++(\arrowangle:\arrowlength);
}
\draw[->,draw=ink,line width=.6pt,>={Stealth[length=1.4mm]}] (16.48,3.48) arc[start angle=-65,end angle=180,radius=.23];
\node[anchor=west] at (16.55,3.78) {$u_i^\vartheta$};
\node at (15.4,1.51) {$a_i\sim\pi_\theta(\,\cdot\mid\mathcal G_t,i)$};

% Independent critic branch reads the global graph embedding.
\draw[flow] (7.2,3.15)--(7.2,.68)--node[above=4pt] {$\mathcal E^V_\phi$} node[below=4pt] {Critic GNN} (9.33,.68);
\foreach \j in {0,...,7}{
  \pgfmathtruncatemacro\shade{25+mod(23*\j,65)}
  \path[draw=white,line width=.3pt,fill=critic!\shade] ({9.58+.13*\j},.575) rectangle ++(.13,.21);
}
\node at (10.10,.22) {$\mathbf g^V$};
\draw[flow] (10.93,.68)--node[above=4pt] {$\mathcal H^V_\phi$} (13.52,.68);
\node[anchor=west] at (13.8,.68) {$V_\phi(\mathcal G_t)$};
\node[anchor=west] at (15.15,.68) {Scene value};
\end{tikzpicture}

%% file: sections/4_experiments.tex
\section{Experiments}

In this section, we evaluate \gls{method} on \bench to study how multiple puzzle pieces can be moved toward their targets in a coordinated manner. In particular, we investigate whether an object-centric multi-agent formulation provides useful structure for this problem and which components are responsible for its performance. Our experiments address three questions: (i) \emph{what architectural structure is needed to learn coordinated multi-object assembly?} (ii) \emph{how do the individual reward terms affect the learned coordination behaviour?} and (iii) \emph{how does the resulting policy compare with classical motion-planning approaches on increasingly constrained assembly tasks?}

\subsection{Experimental Setup}
\label{sec:robot_execution}

We primarily run our experiments and evaluations in simulation. Here we use the MuJoCo implementation of \bench, trained on the GPU with $1024$ instances in parallel. The physics step is $10$ms and the policy acts at $10$Hz. Contacts between pieces are recorded at every physics step. Every episode draws a new target configuration and start poses from the training set. Evaluation uses the held-out configurations. 

\subsection{What Structure Enables Coordinated Assembly?}

First, to understand the effect of each element in our architecture, we isolate the main components of \gls{method} and analyze training and evaluation performance on \benchT. We compare \gls{method} against three architectural variants. {Single-agent PPO} discards both the graph and the multi-agent formulation. Here, an \gls{mlp} maps the concatenated state of all pieces and targets, in a fixed order, to the velocities of all pieces, and is trained using a single clipped objective and the mean per-piece reward. {\gls{mlp}-\gls{mappo}} retains the per-agent formulation of Sec.~\ref{sec:training}, including a shared policy, per-piece advantages, and a centralized critic, but replaces the graph representation with an \gls{mlp} operating on each piece's local state. Finally, {\gls{method}-J} retains the graph actor and critic, but replaces the per-agent \gls{mappo}
objective with a single joint team reward. Fig.~\ref{fig:ablations}-top shows the evolution of task success, the fraction of pieces within $5$cm of their targets, position and yaw errors, and contact rate throughout training for all four architectures, averaged over five seeds. Here, we can see a clear progression as structure is introduced into the policy.  {Single-agent PPO}, which uses neither the graph representation nor the multi-agent decomposition, fails to achieve meaningful task success and retains large position and yaw errors throughout training. Introducing the multi-agent formulation alone already provides a substantial improvement. {\gls{mlp}-\gls{mappo}} brings almost $90\%$ of pieces within $5$cm of their targets, although its overall success plateaus at approximately $50\%$. Adding relational message passing closes this gap, with both graph-based variants, \gls{method} and {\gls{method}-J}, converging to approximately $90\%$ success and substantially lower final position errors. All methods maintain low contact rates, indicating that the differences arise primarily from their ability to coordinate precise placement rather than simply avoid collisions. Interestingly, \gls{method} and {\gls{method}-J} perform very similarly across all metrics, suggesting that while the per-piece advantages of \gls{mappo} gives the policy a useful inductive bias, explicitly modeling relations between pieces through the GNN is essential for success.

\subsection{What Drives Coordinated Behavior?}

\begin{figure*}
\centering
\begin{subfigure}
        \centering
    \includegraphics[width=\linewidth]{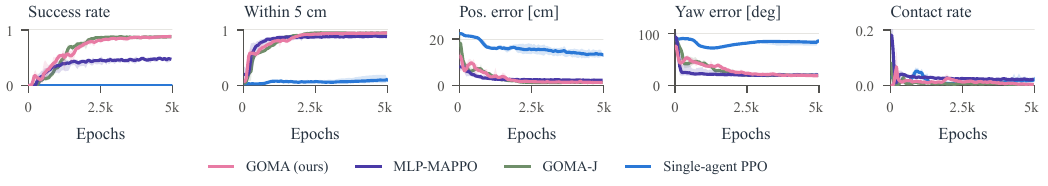}
    \label{fig:3pieceablation}
\end{subfigure}
\begin{subfigure}
    \centering
    \includegraphics[width=\linewidth]{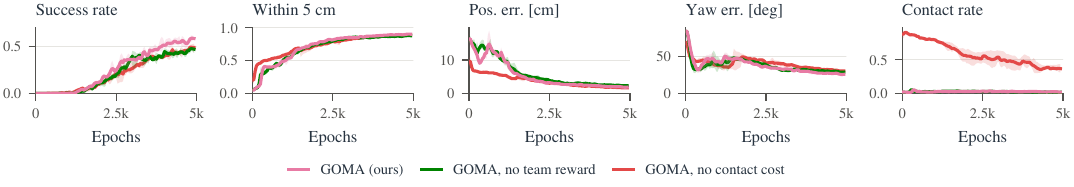}
    \label{fig:reward_ablation}
\end{subfigure}
\vspace{-15pt}
\caption{Policy ablations. On the top row, we show the architecture ablation on \benchT. Here we report task success, fraction of pieces within $5$cm of their targets, position and yaw errors, and contact rate throughout training. On the bottom,  the effect of the reward formulation. We ablate the shared team reward and inter-piece contact cost.}
\label{fig:ablations}
\vspace{-15pt}
\end{figure*}

Next, we study how the reward formulation influences the behavior learned by \gls{method}. We consider two ablations here and evaluate them on \bench-Clusters. In "{\gls{method}, no team reward}", we remove the shared terms of Eq.~\eqref{eq:team_reward}, i.e., the maximum-distance penalty and the global completion bonus, leaving each agent to optimize only its individual reward. In {"\gls{method}, no contact cost}", we remove the inter-piece contact penalty from Eq.~\eqref{eq:individual_reward}. All other reward terms, architectures, and training parameters remain unchanged. In Fig.~\ref{fig:ablations}-bottom we again show the evolution of task success, the fraction of pieces within $5$cm of their targets, position and yaw errors,
and contact rate throughout training, all averaged over five seeds. All three variants learn to move most pieces close to their targets, reaching a similar fraction of pieces within $5$cm and comparable final pose errors. The resulting behaviors, however, are considerably different. Removing the contact cost leads to frequent inter-piece contact throughout training, whereas both variants trained with this penalty reduce contacts to near zero. Removing the shared reward has little effect on collision avoidance or coarse goal reaching, but results in lower final assembly success. This suggests that the individual rewards are sufficient to drive pieces toward their respective targets, while the shared terms help translate this individual progress into completion of the assembly as a whole.

\begin{figure}
    \centering
    \includegraphics[width=\linewidth]{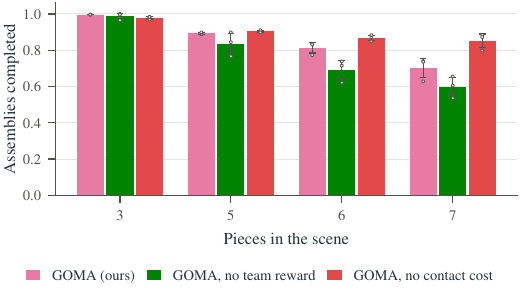}
    \vspace{-15pt}
    \caption{Success comparison on assemblies containing a different piece count.}
    \label{fig:success_scaling}
\end{figure}

We further evaluate these reward variants as the number of pieces increases. Fig.~\ref{fig:success_scaling} compares success on assemblies containing three, five, six, and seven pieces, testing whether the contribution of the shared reward becomes more pronounced as the number of simultaneously interacting pieces grows. We trained on a scene with seven objects and used this trained policy for evaluation of all different piece counts. 

\subsection{How Does Learned Coordination Compare with Classical Planning?}

Finally, we compare \gls{method} against myopic and sampling-based motion planning.
An attractive-only \gls{apf}~\cite{khatib_1986_apf} serves as the myopic baseline.
It drives every piece independently and straight toward its target, using the same tracking controller as the planners.
\emph{Joint \gls{rrt}} ({RRT-J}) plans one path for all pieces at once with RRT-Connect~\cite{kuffner2000rrtconnect} in the $21$-dimensional configuration space and with mesh collision checks.
\emph{Per-piece \gls{rrt}} ({RRT-P}) plans a $3$-dimensional path for every piece against the start poses of the others and executes all paths at once.
Both planners are also evaluated with a replan mode, where a new plan is generated every $5$s. This is added so that they can react to the movements of other pieces.
Under one-piece-at-a-time execution, the joint plan is followed piece by piece, and the per-piece paths are planned in sequence with the earlier pieces assumed at their goals.
All methods receive the same target assignments and the same initial configurations. The planners are given the $117$ of $258$ held-out instances whose start arrangement passes their collision check and a $150$\,s episode, so that planning time does not decide the outcome.
In Tab.~\ref{tab:planning_comparison} we compare the methods on \benchC and \benchP.
We report assembly success, final position error, the fraction of control steps with inter-piece contact, the median time to completion, and the computation time at inference.

These results highlight the importance of coordination. Crucially, \gls{apf} and {RRT-P} assemble at most $9\%$ of the clusters and none of the puzzles, with inter-piece contact during nearly every control step.
Joint planning with replanning achieves the highest success under concurrent execution ($98\%$ on \benchC and $88\%$ on \benchP), compared with $75\%$ and $41\%$ for \gls{method}.
However, its contact rates are also higher. $12\%$ and $37\%$, respectively, compared with $4\%$ for \gls{method} on both benchmarks.
Additionally, \gls{method} requires approximately $0.68$ms per inference and achieves median completion times of $8$\,s on \bench-Cluster and $24$\,s on \bench-Puzzle, compared with $21$\,s and $43$s for RRT-J with replanning.
Thus, \gls{method} offers fast, low-contact execution, while joint planning with replanning remains more reliable at completing tightly constrained assemblies.

\input{sections/table_classical}

\subsection{Zero-shot Sim2real transfer}

To demonstrate the applicability of \gls{method} beyond simulation, we perform a qualitative zero-shot sim-to-real demonstration on a Franka FR3 manipulator, without additional policy training or fine-tuning. The robot is controlled using the CRISP Cartesian impedance controller and is equipped with the custom square-peg manipulation tool introduced in Section~\ref{sec:bench}. Piece poses are estimated using a hand--eye calibrated camera setup consisting of a scene-mounted Intel RealSense D435i RGB-D camera for global pose estimation and a wrist-mounted RealSense D405 for close-range localization of the insertion socket. We estimate the pose of each piece using FoundationPose~\cite{Wen_2024_CVPR}, with HSV color segmentation providing an individual mask for each of the seven pieces. Known geometric offsets transform the estimated model poses to the corresponding piece centroids, while a
geometric consistency filter rejects estimates that would imply physically impossible overlaps.

Since the physical robot can manipulate only one piece at a time, we execute the multi-agent policy sequentially. A round-robin strategy selects among unfinished pieces, allowing a piece to remain temporarily in place and be revisited in subsequent rounds. For each selected piece, the wrist-mounted camera first localizes its insertion socket, after which the robot engages the tool and executes the motion predicted by the corresponding agent. During execution, the pose of the manipulated piece is updated from the robot's forward
kinematics, while the poses of the remaining pieces are obtained from perception. The resulting object poses are used to reconstruct the interaction graph before each policy query. Fig.~\ref{fig:hero} shows the resulting real-world assembly sequence, illustrating that the policy can be deployed on the physical setup without additional real-world training.

%% file: sections/table_classical.tex
\begin{table}[t]
  \centering
  \caption{\gls{method} against the \gls{apf} controller and joint (RRT-J) and per-piece (RRT-P) RRT planning, planned once or replanned every $5$\,s, on the held-out \benchC and \benchP.}
  \label{tab:planning_comparison}
  \renewcommand{\arraystretch}{1.1}
  \begin{tabular*}{\columnwidth}{@{\extracolsep{\fill}}lrrrrr@{}}
  \toprule
  Method & Succ. & Err. & Contact & Time & Compute \\
         & (\%) & (cm) & (\%) & (s) & \\
  \midrule
  \multicolumn{6}{@{}l}{\emph{\benchC}} \\
  \gls{method} (ours) & {75} & {1.0} & {4} & {8} & {0.68\,ms} \\
  APF & 3 & 3.8 & 99 & 80 & $\approx 0$ \\
  RRT-J & 73 & 1.9 & 44 & 20 & 774\,ms \\
  RRT-J +replan & 98 & 0.3 & 12 & 21 & 49\,ms \\
  RRT-P & 3 & 4.0 & 98 & 119 & 42\,ms \\
  RRT-P +replan & 9 & 3.2 & 95 & 85 & 6.7\,ms \\
  \midrule
  \multicolumn{6}{@{}l}{\emph{\benchP}} \\
  \gls{method} (ours) & {41} & {1.7} & {4} & {24} & {0.68\,ms} \\
  APF & 0 & 4.3 & 100 & -- & $\approx 0$ \\
  RRT-J & 58 & 3.3 & 59 & 24 & 6.8\,s \\
  RRT-J +replan & 88 & 1.5 & 37 & 43 & 1.1\,s \\
  RRT-P & 0 & 4.5 & 99 & -- & 48\,ms \\
  RRT-P +replan & 0 & 4.1 & 99 & -- & 0.20\,ms \\
  \bottomrule
  \end{tabular*}
\end{table}

%% file: sections/5_conclusion.tex
\section{Conclusion}
In this work, we studied planar geometric assembly as a coordinated multi-object manipulation problem. We introduced \bench, a procedural Tangram benchmark with graded levels of geometric coupling, together with simulation and a corresponding real-robot setup. On this benchmark, we formulated assembly from an object-centric multi-agent perspective and proposed \gls{method}, where each piece is represented as an agent and relational information is exchanged through a graph neural network. Our experiments investigate which components of this formulation enable coordinated behavior and compare the resulting policy with classical motion-planning approaches across increasingly constrained assembly configurations.

Our formulation intentionally isolates coordination from several complementary challenges in robotic assembly. In particular, the target pose and piece-to-target assignments are provided. We therefore do not address the problem of inferring a Tangram decomposition from a target silhouette and determining which piece belongs to each region. Incorporating these capabilities would connect the proposed benchmark more directly to the complete Tangram-solving problem and represents an interesting direction for future work. Similarly, while our simulated formulation permits all pieces to move concurrently, the current physical setup uses a single robot and therefore executes the multi-agent policy sequentially. Extending the benchmark to bimanual or multi-robot systems would allow the concurrent coordination learned by the policy to be evaluated directly in the physical world. More broadly, \bench abstracts away traditional nonprehensile pushing and constrains manipulation to planar motion through the insertion tool, allowing us to focus specifically on the interactions between pieces. Future work could relax these assumptions by incorporating pushing, richer contact dynamics, partial observability, and uncertainty in perception, as well as extending the procedural generation framework beyond Tangram to more general planar assemblies and ultimately full 3D assembly problems.

%% file: sections/6_appendix.tex
% \par\noindent\begin{minipage}{\columnwidth}
\appendix[Algorithm Hyperparameters]
\label{sec:hyperparameters}
\centering
\captionof{table}{Hyperparameters for \glsentryshort{method}.}
\label{tab:hyperparameters}
\small
\setlength{\tabcolsep}{4pt}
\renewcommand{\arraystretch}{1.08}
\begin{tabularx}{\columnwidth}{@{}Xr@{}}
\toprule
\textbf{Algorithm setting} & \textbf{Value} \\
\midrule
Latent and hidden dimension $H$ & $256$ \\
Message-passing rounds $K$ & $1$ \\
Node embedding and update blocks & $9$ \\
Edge \& global embedding \& update blocks & $8$ \\
Policy and critic head blocks & $1$ \\
Policy learning rate & $2\!\times\!10^{-5}\rightarrow10^{-6}$ \\
Critic learning rate & $6\!\times\!10^{-5}\rightarrow3\!\times\!10^{-6}$ \\
Discount factor $\gamma$ & $0.99$ \\
Advantage parameter $\lambda$ & $0.9$ \\
Policy and value clipping $\eta$ & $0.1$ \\
Update epochs & $3$ \\
Minibatch size & $4096$ \\
Entropy coefficient & $0.0$ \\
Maximum gradient norm & $5$ \\
Rollout length & $100$ \\
Parallel environments & $1024$ \\
\bottomrule
\end{tabularx}
% \end{minipage}\par